\documentclass{article}

\usepackage[english]{babel}
\usepackage[utf8x]{inputenc}
\usepackage[T1]{fontenc}

\usepackage{amsmath}
\usepackage{amssymb}

\newcommand{\bma}{\bm{a}}
\newcommand{\bmb}{\bm{b}}

\newcommand{\bmg}{\bm{g}}
\newcommand{\bmh}{\bm{h}}

\newcommand{\bmk}{\bm{k}}

\newcommand{\bmq}{\bm{q}}

\newcommand{\bmy}{\bm{y}}

\newcommand{\bmA}{\bm{A}}

\newcommand{\bmG}{\bm{G}}
\newcommand{\bmH}{\bm{H}}

\newcommand{\bmK}{\bm{K}}

\newcommand{\bmQ}{\bm{Q}}

\newcommand{\bmU}{\bm{U}}
\newcommand{\bmV}{\bm{V}}
\newcommand{\bmW}{\bm{W}}
\newcommand{\bmX}{\bm{X}}

\usepackage{dutchcal}

\newcommand{\mcL}{\mathcal{L}}

\newcommand{\mcN}{\mathcal{N}}

\newcommand{\mcR}{\mathcal{R}}

\newcommand{\mcY}{\mathcal{Y}}

\newcommand{\mbbR}{\mathbb{R}}

\DeclareMathOperator*{\softmax}{softmax}

\DeclareMathOperator*{\leakyrelu}{LeakyReLu}

\newcommand{\ith}{$i^{\text{th}}$}

\usepackage{tikz}

\usetikzlibrary{fit,positioning,arrows.meta}
\tikzset{neuron/.style={shape=circle, minimum size=1.25cm, 
    inner sep=0, draw, font=\small}, io/.style={neuron, fill=gray!20}}

\usepackage{iclr_2019}

\usepackage[colorinlistoftodos]{todonotes}

\usepackage{amsthm}
\usepackage{amsfonts}

\usepackage{bm}
\usepackage{graphicx}
\usepackage{hyperref}
\hypersetup{
    colorlinks = true,
    citecolor = {olive}
}
\usepackage[acronym,smallcaps,nowarn,section,nonumberlist]{glossaries}
\usepackage{algorithm}
\usepackage{algorithmicx}
\usepackage[noend]{algpseudocode}
\makeglossaries

\newacronym{argat}{ARGAT}{Across-Relation Graph Attention}
\newacronym{auc}{AUC}{Area Under the Curve}

\newacronym{bow}{BOW}{bag-of-words}
\newacronym{bn}{BN}{Bayesian Network}

\newacronym{cdf}{CDF}{Cumulative Distribution Function}
\newacronym{cnn}{CNN}{Convolutional Neural Network}
\newacronym{cvae}{CVAE}{Conditional Variational Autoencoder}
\newacronym{dag}{DAG}{Directed Acyclic Graph}
\newacronym{dgm}{DGM}{Directed Graphical Model}
\newacronym{dnn}{DNN}{Deep Neural Network}
\newacronym{ecfp4}{ECFP4}{Extended Connectivity Fingerprint}

\newacronym{gat}{GAT}{Graph Attention Network}
\newacronym{gcn}{GCN}{Graph Convolutional Network}
\newacronym{gdl}{GDL}{Geometric Deep Learning}
\newacronym{ggnn}{GGNN}{Gated Graph Neural Network}
\newacronym{glove}{GloVe}{Global Vectors for Word Representation}
\newacronym{gnn}{GNN}{Graph Neural Network}
\newacronym{gru}{GRU}{Gated Recurrent Unit}

\newacronym{hmm}{HMM}{Hidden Markov Model}
\newacronym{kld}{KLD}{Kullback-Leibler Divergence}
\newacronym{knn}{KNN}{K-Nearest Neighbour}
\newacronym{idf}{IDF}{Inverse Document Frequency}
\newacronym{ig}{IG}{Information Gain}
\newacronym{isan}{ISAN}{Input Switched Affine Network}
\newacronym{lhs}{L.H.S.}{left hand side}
\newacronym{lstm}{LSTM}{Long Short Term Memory}

\newacronym{map}{MAP}{Maximum a Posteriori}
\newacronym{mdl}{MDL}{Minimum Description Length}
\newacronym{ml}{ML}{Machine Learning}
\newacronym{mle}{MLE}{Maximum Likelihood Estimator}
\newacronym{mnist}{MNIST}{Modified National Institute of Standards and Technology}

\newacronym{nlp}{NLP}{Natural Language Processing}
\newacronym{nn}{NN}{Neural Network}

\newacronym{oh}{OH}{One-Hot}

\newacronym{pdf}{PDF}{Probability Density Function}
\newacronym{pgm}{PGM}{Probabilistic Graphical Model}

\newacronym{relu}{ReLU}{Rectified Linear Unit}
\newacronym{rdf}{RDF}{Resource Description Framework}
\newacronym{rgb}{RGB}{Red Green Blue}
\newacronym{rgc}{RGC}{Relational Graph Convolution}
\newacronym{rgat}{RGAT}{Relational Graph Attention Network}
\newacronym{rgcn}{RGCN}{Relational Graph Convolutional Network}
\newacronym{rhs}{R.H.S.}{Right Hand Side}
\newacronym{rnn}{RNN}{Recurrent Neural Network}
\newacronym{roc}{ROC}{Receiver Operating Characteristic}

\newacronym{sgd}{SGD}{Stochastic Gradient Descent}
\newacronym{sg}{SG}{SkipGram}
\newacronym{st}{ST}{SkipThought}
\newacronym{sts}{STS}{Semantic Textual Similarity}

\newacronym{termfreq}{TF}{Term Frequency}
\newacronym{tfidf}{TF-IDF}{Term Frequency-Inverse Document Frequency}

\newacronym{vae}{VAE}{Variational Autoencoder}

\newacronym{wirgat}{WIRGAT}{Within-Relation Graph Attention}
\newacronym{wl}{WL}{Weisfeiler-Lehman}
\usepackage{tikz}

\usetikzlibrary{fit,positioning,arrows.meta}
\tikzset{neuron/.style={shape=circle, minimum size=1.25cm, 
    inner sep=0, draw, font=\small}, io/.style={neuron, fill=gray!20}}
\glsdisablehyper
\usepackage{cleveref}

\usepackage{url}            
\usepackage{booktabs}       
\usepackage{nicefrac}       
\usepackage{microtype}      
\usepackage{subcaption}

\theoremstyle{definition}

\crefname{lemma}{lemma}{lemmas}
\Crefname{lemma}{Lemma}{Lemmas}

\crefname{lemma}{lemma}{lemmas}
\Crefname{lemma}{Lemma}{Lemmas}

\title{Relational Graph Attention Networks}

\author{Dan Busbridge, Dane Sherburn, Pietro Cavallo \& Nils Y. Hammerla\\
Babylon Health\\
60 Sloane Avenue\\
SW3 3DD\\
London, United Kingdom\\
\texttt{\{dan.busbridge, dane.sherburn, pietro.cavallo}\\\texttt{\phantom{\{}nils.hammerla\}@babylonhealth.com}
}

\iclrfinalcopy

\begin{document}
\maketitle

\begin{abstract}
We investigate Relational Graph Attention Networks, a class of models that extends
non-relational graph attention mechanisms to incorporate
relational information, opening up these methods to a wider variety of problems.
A thorough evaluation of these models is performed, and comparisons are made
against established benchmarks. To provide a meaningful comparison, we retrain Relational Graph
Convolutional Networks, the spectral counterpart of Relational Graph Attention
Networks, and evaluate them under the same conditions.
We find that Relational Graph Attention Networks perform worse
than anticipated, although some configurations are
marginally beneficial for modelling molecular properties.
We provide insights as to why this may be, and suggest both modifications
to evaluation strategies, as well as directions to investigate for future work.
\end{abstract}
\section{Introduction}
\label{sec:introduction}
\glsresetall

\glspl{cnn} successfully solve a variety of tasks in Euclidean grid-like domains, such as image captioning \citep{Donahue2017} and
classifying videos
\citep{Karpathy2014}. \glspl{cnn} are successful because they assume the data
is locally stationary and compositional \citep{Defferrard2016, Henaff,
  Bruna}.

However, data often occurs in the form of graphs or manifolds, which are classic examples of non-Euclidean domains. Specific instances include knowledge bases, molecules, and point clouds captured by 3D data acquisition devices \citep{Wang2018}. The generalisation of \glspl{nn} to non-Euclidean domains is termed \gls{gdl}, and may be roughly divided into spectral, spatial and hybrid approaches \citep{Bronstein}.

Spectral approaches \citep{Defferrard2016}, most notably \glspl{gcn} \citep{Kipf2016}, are limited by their basis-dependence. A filter that is learned with respect to a basis on one domain is not guaranteed to behave similarly when applied to another basis and domain. Spatial approaches are limited by an absence of shift invariance and lack of coordinate system \citep{Duvenaud,Atwood2016,Monti}. Hybrid approaches combine spectral and spatial approaches, trading their advantages and deficiencies against each-other \citep{Bronstein,Rustamov2013,Szlam,Gavish2010}.

A recent approach that began with \glspl{gat}, applied attention mechanisms to graphs, and does not share these
limitations
\citep{Velickovic2017,Gong2018a,Zhang2018b,Monti2018a,Lee2018}.

An alternative direction has been to generalise \glspl{rnn} from sequential message passing on one-dimensional signals, to message passing on graphs \citep{sperduti1997,Frasconi97onthe,Gori2005}. Incorporating gating mechanisms led to the development of \glspl{ggnn} \citep{Scarselli2009,allamanis2017}\footnote{We note that \glspl{ggnn} support relation types. Evaluating these models on the tasks presented here is necessary to acquire a better understanding neural models of relational data.}.

\glspl{rgcn} have been proposed as an extension of \glspl{gcn} to the domain of
relational graphs \citep{Schlichtkrull2018}. This model has achieved
impressive performance on node classification and link prediction tasks, however, its mechanisms still resides within spectral methods and shares their deficiencies. The focus of this work investigate generalisations of \gls{rgcn} away from its spectral origins.

We take \gls{rgcn} as a starting point, and investigate a class of models we term
\glspl{rgat}, extending attention mechanisms to the relational graph domain.
We consider two variants, \gls{wirgat} and \gls{argat}, each with either
additive or multiplicative attention. We perform an extensive hyperparameter
search, and evaluate these models on challenging transductive node
classification and inductive graph classification tasks. These models are
compared against established benchmarks, as well as a re-tuned \gls{rgcn} model.

We show that \gls{rgat} performs worse than expected, although some
configurations produce marginal benefits on
inductive graph classification tasks. In order to aid further investigation in
this direction, we present the full \glspl{cdf} for the
hyperparameter searches in \Cref{app:cdfs}, and statistical hypothesis tests in
\Cref{app:significance}. We also provide a vectorised, sparse, batched
implementation of \gls{rgat} and \gls{rgcn} in \texttt{TensorFlow} which is
compatible with \texttt{eager} execution mode to open up research into these models to a wider audience\footnote{\url{https://github.com/Babylonpartners/rgat}.}.

\section{RGAT architecture}
\label{sec:rgat-architecture}
\subsection{Relational graph attention layer}
\label{subsec:rgat-architecture/relational-graph-attention-layer}

We follow the construction of the \gls{gat} layer in \cite{Velickovic2017},
extending to the relational setting, using ideas from \cite{Schlichtkrull2018}.

\paragraph{Layer input and output}
The input to the layer is a graph
with $R=|\mcR|$ relation types and $N$ nodes.
The \ith \, node is represented by a feature vector $\bmh_i\in\mbbR^F$, and the features of
all nodes are summarised in the feature matrix
$\bmH=[\bmh_1 \, \bmh_2 \,\ldots \, \bmh_N]\in\mbbR^{N\times F}$.
The output of the layer is the transformed feature matrix
$\bmH^\prime=[\bmh_1^\prime \, \bmh_2^\prime \,\ldots \,
\bmh_N^\prime]\in\mbbR^{N\times F^\prime}$,
where
$\bmh_i^\prime\in\mbbR^{F^\prime}$
is the 
transformed feature vector of the \ith \, node.

\paragraph{Intermediate representations}
Different relations convey distinct pieces of information.
The update rule of \cite{Schlichtkrull2018} made this manifest by assigning each
node $i$ a distinct intermediate representation
$\bmg_i^{(r)}\in\mbbR^{F^\prime}$ under relation $r$
\begin{equation}
  \label{eq:rgat-linear}
  \bmG^{(r)}=
  \bmH \, \bmW^{(r)} \in \mbbR^{N\times F^\prime},
\end{equation}
where $\bmG^{(r)}=\left[\bmg_1^{(r)}\,\bmg_2^{(r)}\,\ldots\,\bmg_N^{(r)}\right]$
is the intermediate representation feature matrix under relation $r$, and
$\bmW^{(r)}\in\mbbR^{F\times F^\prime}$ are the learnable parameters of a shared
linear transformation.

\paragraph{Logits}
Following \cite{Velickovic2017,Zhang2018b}, we assume the attention coefficient between two
nodes is based only on the features of those nodes up to a
neighborhood-level normalisation.
To keep computational complexity
linear in $R$, we assume that, given linear transformations $\bmW^{(r)}$, the
logits $E_{i,j}^{(r)}$ of each relation $r$ are independent
\begin{equation}
  \label{eq:relational-logits-general}
  E_{i,j}^{(r)} = a\left(\bmg_i^{(r)}, \bmg_{j}^{(r)}\right),
\end{equation}
and indicate the importance of node $j$'s intermediate representation to that of
node $i$ under relation $r$. The attention is masked so that, for node $i$,
coefficients $\alpha_{i,j}^{(r)}$ exist only for $j\in\mcN_i^{(r)}$, where 
$\mcN_i^{(r)}$ denotes the set of neighbor indices of node $i$ under
relation $r\in\mcR$.

\begin{figure}[t]
  \begin{center}
    \includegraphics[width=0.8\linewidth]{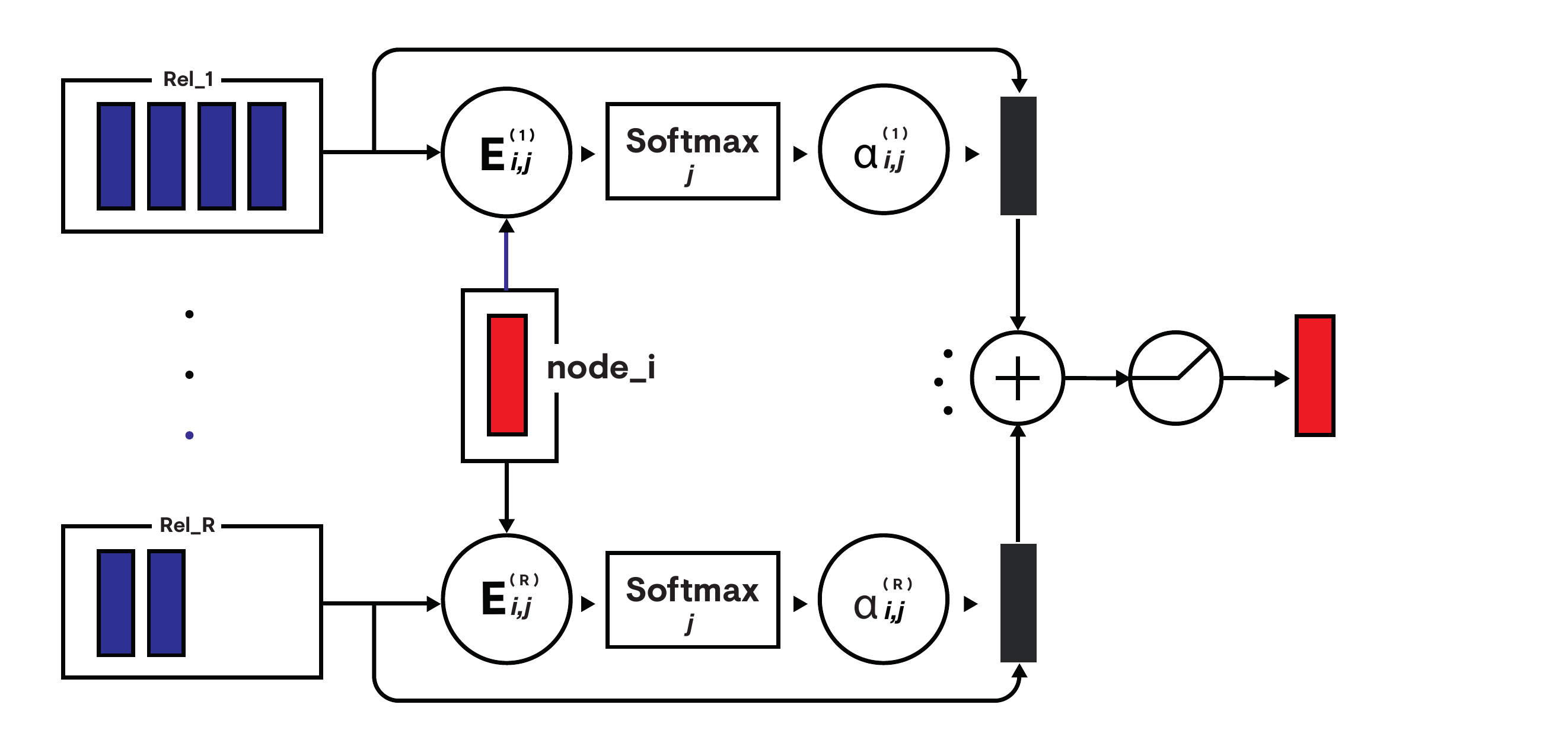}
  \end{center}
  \caption{\gls{wirgat}. The intermediate representations for node $i$ (left red
    rectangle) are combined with the intermediate representations for nodes in
    its neighborhood (blue rectangles) under each relation $r$, to form each
    logit $E_{i,j}^{(r)}$. A softmax is taken over each logit matrix
    for each relation type to form
    the attention coefficients $\alpha_{i,j}^{(r)}$. These attention
    coefficients construct a weighted sum over the nodes in the neighborhood for
    each relation (black rectangle). These are then aggregated and passed
    through a nonlinearity to produce the updated representation for node $i$
    (right red rectangle).}
  \label{fig:wirgat}
\end{figure}

\paragraph{Queries, keys and values}
The logits are composed from queries and keys, and
specify how the values, i.e. the intermediate representations $\bmg_i^{(r)}$,
will combine to produce the updated node representations $\bmh_i^\prime$ \citep{Vaswani2017}.
A separate query kernel
$\bmQ^{(r)}\in\mbbR^{F^\prime\times D}$
and key kernel
$\bmK^{(r)}\in\mbbR^{F^\prime\times D}$
project the intermediate representations
$\bmg_i^{(r)}$,
into query and key representations of dimensionality $D$
\begin{align}
  \bmq_i^{(r)} 
  &= \bmg_i^{(r)} \, \bmQ^{(r)}
  \in\mbbR^{D},
  &
  \bmk_i^{(r)} 
  &= \bmg_i^{(r)} \, \bmK^{(r)}
  \in\mbbR^{D}.
\end{align}
For convenience, the query and key kernels are combined to form the
attention kernels
$\bmA^{(r)}=\bmQ^{(r)}\oplus\bmK^{(r)}\in\mbbR^{2F^\prime\times D}$.
These query and key representations are the building blocks of
the two specific realisations of $a$ in \Cref{eq:relational-logits-general}
that we now consider.
\paragraph{Additive attention logits} 
The first realisation of $a$ we consider is the relational modification of the
logit mechanism of
\cite{Velickovic2017}
\begin{equation}
  \label{eq:relational-logits-specific}
  E_{i,j}^{(r)}
  = \leakyrelu\left( q_i^{(r)}+k_j^{(r)} \right),
\end{equation}
where the query and key dimensionality are both $D=1$, and $q_i^{(r)}$ and $k_i^{(r)}$
are scalar flattenings of their one-dimensional vector counterparts
$\bmq_i^{(r)},\bmk_i^{(r)}\in\mbbR^{1}$.
We refer to any instance of \gls{rgat} using logits of the form in
\Cref{eq:relational-logits-specific} as additive \gls{rgat}.
\paragraph{Multiplicative attention logits}
The second realisation we consider is the multiplicative mechanism of
\cite{Vaswani2017,Zhang2018b}\footnote{The form of our mechanism is not
  precisely that of \cite{Zhang2018b} as they also consider residual
  concatenation and gating mechanism
  applied across the heads of the attention mechanism.}
\begin{equation}
  \label{eq:relational-dot-logits-specific}
  E_{i,j}^{(r)}
  = \bmq_i^{(r)}\cdot \bmk_j^{(r)},
\end{equation}
where the query and key dimensionality $D$ can be any positive integer.
We refer to any instance of \gls{rgat} using logits of the form in
\Cref{eq:relational-logits-specific} as multiplicative \gls{rgat}.

It should be noted that there are many types of attention mechanisms beyond
vanilla additive and multiplicative. These include mechanisms leveraging the structure
of the dual graph \cite{Monti2018a} as well as learned edge features
\cite{Gong2018a}.

The attention coefficients should be comparable across nodes. This can be achieved
by applying softmax appropriately to any logits $E_{i,j}^{(r)}$.
We investigate two candidates, each encoding a different prior belief about how the importance of different relations.
\paragraph{\gls{wirgat}}
The simplest way to take the softmax over the logits $E_{i,j}^{(r)}$ of
\Cref{eq:relational-logits-specific} or
\Cref{eq:relational-dot-logits-specific} is to do so independently for each relation $r$
\begin{align}
  \label{eq:wirgat-softmax}
  \alpha_{i,j}^{(r)}
  &=\softmax_{j}\left( E_{i,j}^{(r)} \right)
    =\frac{\exp\left( E_{i,j}^{(r)} \right)}{\sum_{k\in\mcN_i^{(r)}}\exp\left( E_{i,k}^{(r)} \right)},
  &
   \forall \, i,r: \sum_{j\in\mcN_i^{(r)}} \alpha_{i,j}^{(r)}=1.
\end{align}
We call the attention in
\Cref{eq:wirgat-softmax} \glsfirst{wirgat} (\gls{wirgat}), and it is shown in \Cref{fig:wirgat}.
This mechanism encodes the prior that relation importance is a purely
global property of the graph by implementing an independent probability
distribution over nodes in the neighborhood of $i$ for each relation $r$. 
Explicitly, for any
node $i$ and relation $r$, nodes $j,k\in\mcN_i^{(r)}$ yield competing attention
coefficients $\alpha_{i,j}^{(r)}$ and $\alpha_{i,k}^{(r)}$ with sizes
depending on their corresponding representations
$\bmg_{j}^{(r)}$ and $\bmg_{k}^{(r)}$. There is no competition
between any attention coefficients $\alpha_{i,j}^{(r)}$ and
$\alpha_{i,k}^{(r^\prime)}$for all nodes $i$ and nodes $j\in\mcN_i^{(r)},
j^\prime\in\mcN^{(r^\prime)}$ where $r^\prime\neq r$ irrespective of node
representations.

\begin{figure}[t]
  \begin{center}
    \includegraphics[width=0.8\linewidth]{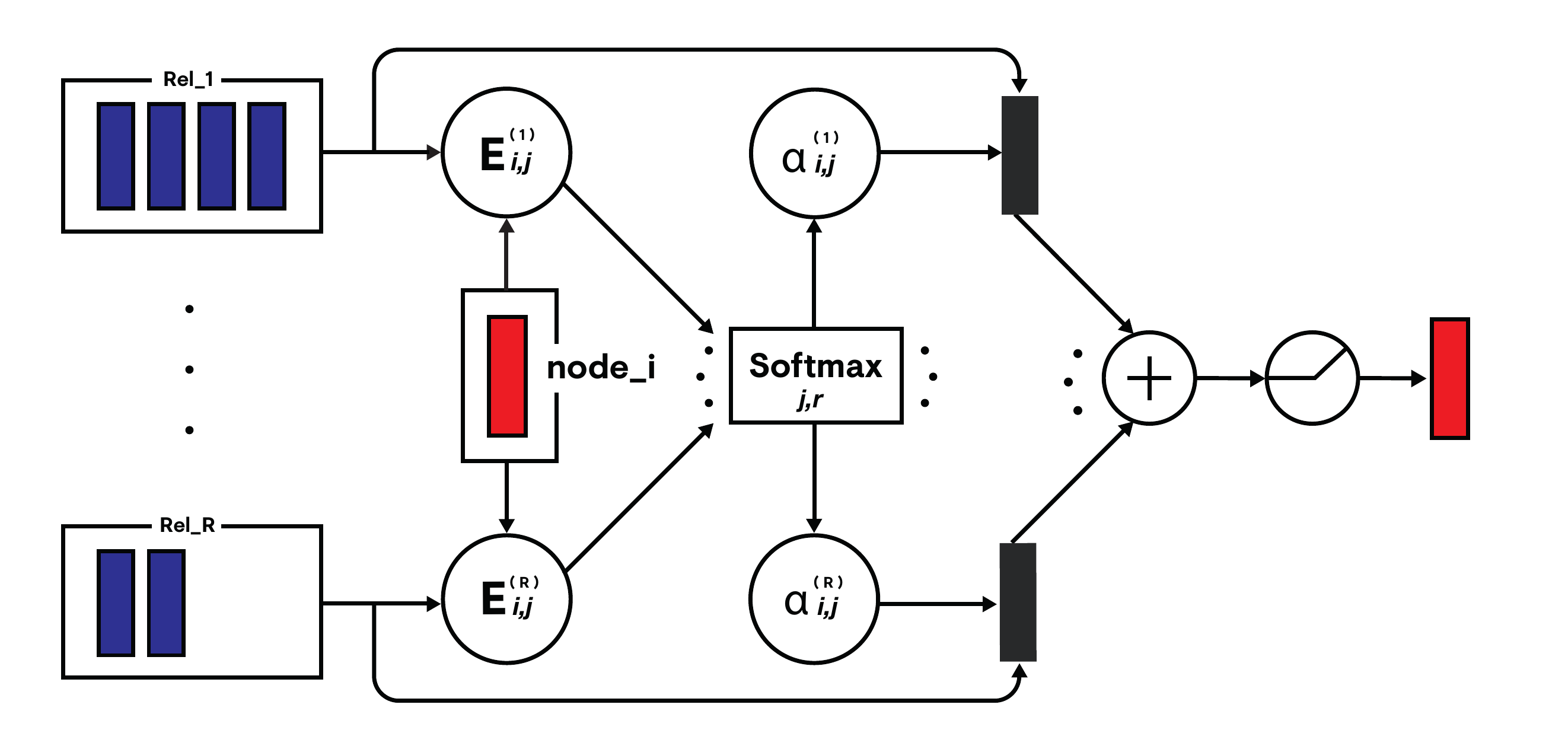}
  \end{center}
  \caption{\gls{argat}. The logits are produced identically to those in
    \Cref{fig:wirgat}. A softmax is taken across all logits independent of
    relation type to form the attention coefficients $\alpha_{i,j}^{(r)}$. The
    remaining weighting and aggregation steps are the same as those in
    \Cref{fig:wirgat}.}
  \label{fig:argat}
\end{figure}

\paragraph{\gls{argat}} An alternative way to take the
softmax over the logits $E_{i,j}^{(r)}$ of
\Cref{eq:relational-logits-specific} or
\Cref{eq:relational-dot-logits-specific}
is across node neighborhoods irrespective of relation $r$
\begin{align}
  \label{eq:argat-softmax}
  \alpha_{i,j}^{(r)}
  &=\softmax_{j,r}\left( E_{i,j}^{(r)} \right)
  =\frac{
    \exp\left( E_{i,j}^{(r)} \right)}{
    \sum_{r^\prime\in\mcR}\sum_{k\in\mcN_i^{(r^\prime)}}\exp\left( E_{i,k}^{(r^\prime)} \right)},
  &
    \forall\,i:\sum_{r\in\mcR}\sum_{j\in\mcN_i^{(r)}} \alpha_{i,j}^{(r)}=1.
\end{align}
We call the attention in
\Cref{eq:argat-softmax} \Cref{eq:wirgat-softmax} \glsfirst{argat} (\gls{argat}),
and it is shown in \Cref{fig:argat}.
This mechanism encodes the prior that
relation importance is a local property of the graph by implementing a
single probability distribution over the different representations
$\bmg_j^{(r)}$ for nodes j in the neighborhood of node $i$.
Explicitly, for any
node $i$ and all $r,r^\prime\in\mcR$, all nodes $j\in\mcN_i^{(r)}$ and
$k\in\mcN_i^{(r^\prime)}$
yield competing attention
coefficients $\alpha_{i,j}^{(r)}$ and $\alpha_{i,k}^{(r^\prime)}$ with sizes
depending on their corresponding representations
$\bmg_{j}^{(r)}$ and $\bmg_{k}^{(r^\prime)}$.

\paragraph{Comparison to \gls{rgcn}}
For comparison, the coefficients of
\gls{rgcn} are given by
$\alpha_{i,j}^{(r)} = |\mcN_i^{(r)}|^{-1}$.
This encodes the prior that the intermediate
representations of nodes $j\in\mcN_i^{(r)}$ to node $i$ under relation $r$ are
equally important.

\paragraph{Propagation rule}
Combining the attention mechanism of either 
\Cref{eq:wirgat-softmax} or \Cref{eq:argat-softmax} with the neighborhood
aggregation step of \cite{Schlichtkrull2018} gives 
\begin{equation}
  \bmh_i^\prime = \sigma\left(
    \sum_{r\in\mcR}
    \sum_{j\in\mcN_i^{(r)}}
    \alpha_{i,j}^{(r)}\,\bmg_j^{(r)}
  \right)\in\mbbR^{N\times F^\prime},
\end{equation}
where $\sigma$ represents an optional nonlinearity. Similar to
\cite{Vaswani2017,Velickovic2017}, we also find that using multiple heads in the
attention mechanism can enhance performance
\begin{equation}
  \bmh_i^\prime = \bigoplus_{k=1}^K \sigma\left(
    \sum_{r\in\mcR}
    \sum_{j\in\mcN_i^{(r)}}
    \alpha_{i,j}^{(r, k)}\,\bmg_j^{(r, k)}
  \right)\in\mbbR^{N\times K\,F^\prime},
\end{equation}
where $\oplus$ denotes vector concatenation, $\alpha_{i,j}^{(r,k)}$ are the
normalised attention coefficients under relation $r$ computed by either
\gls{wirgat} or \gls{argat}, and $\bmg_i^{(r,k)} =
\bmh_i\,\left(\bmW^{(r,k)}\right)^T$ is the head specific intermediate
representation of node $i$ under relation $r$.

It might be interesting to
consider cases where there are a different number of heads for different
relationship types, as well as when a mixture of \gls{argat} and \gls{wirgat}
produce the attention coefficients, however, we leave that subject for future
investigation and will not consider it further.

\paragraph{Basis decomposition}
The number of parameters in the \gls{rgat} layer increases linearly with the
number of relations $R$ and heads $K$, and can lead quickly to
overparameterisation. In \glspl{rgcn} it was found that decomposing the kernels
was beneficial for generalisation, although it comes at the cost of increased
model bias \citep{Schlichtkrull2018}. We follow this approach, decomposing both
the kernels $\bmW^{(r,k)}$ as well as the kernels of attention mechanism
$\bmA^{(r,k)}$ into $B_V$ basis matrices $\bmV^{(b)}\in\mbbR^{F\times
  F^\prime}$ and $B_X$ basis vectors $\bmX^{(b)}\in\mbbR^{2F^\prime\times D}$
\begin{align}
  \label{eq:w-a-decomposition}
  \bmW^{(r,k)}
  &= \sum_{b=1}^{B_W} c^{(r,k)}_{b}\,\bmV^{(b)},
  & \bmA^{(r,k)}
  &= \sum_{b=1}^{B_X} d^{(r,k)}_b\,\bmX^{(b)},
\end{align}
where $c^{(r,k)}_b,d^{(r,k)}_b\in\mbbR$ are basis coefficients. We consider
models using full and decomposed $\bmW$ and $\bmA$.

\subsection{Node classification}
\label{sec:node-classification}

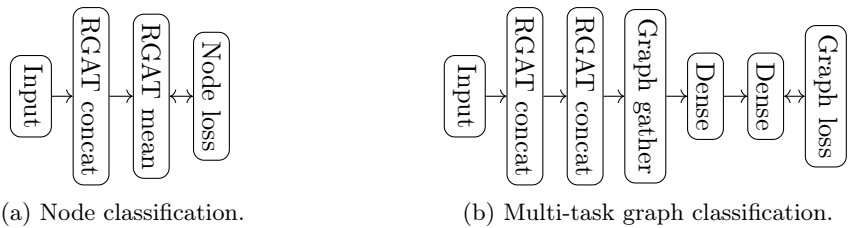
\begin{figure}
\begin{subfigure}{.5\textwidth}
  \centering
  \begin{tikzpicture}
  \tikzset{layer/.style={rectangle,draw,rounded corners,rotate=-90}}
  \node[layer] (a) at (0,0) {Input};
  \node[layer] (b) at (0.8,0) {RGAT concat};
  \node[layer] (c) at (1.6,0) {RGAT mean};
  \node[layer] (d) at (2.4,0) {Node loss};
  \draw[->]  (a) edge (b) ;
  \draw[->]  (b) edge (c) ;
  \draw[<->] (c) edge (d) ;
\end{tikzpicture}
  \caption{Node classification.} 
  \label{sfig:architecture-node-classification}
\end{subfigure}%
\begin{subfigure}{.5\textwidth}
  \centering
  \begin{tikzpicture}
  \tikzset{layer/.style={rectangle,draw,rounded corners,rotate=-90}}
  \node[layer] (a) at (0,0) {Input};
  \node[layer] (b) at (0.8,0) {RGAT concat};
  \node[layer] (c) at (1.6,0) {RGAT concat};
  \node[layer] (d) at (2.4,0) {Graph gather};
  \node[layer] (e) at (3.2,0) {Dense};
  \node[layer] (f) at (4,0) {Dense};
  \node[layer] (g) at (4.8,0) {Graph loss};
  \draw[->]  (a) edge (b) ;
  \draw[->]  (b) edge (c) ;
  \draw[->]  (c) edge (d) ;
  \draw[->]  (d) edge (e) ;
  \draw[->]  (e) edge (f) ;
  \draw[<->] (f) edge (g) ;
\end{tikzpicture}
  \caption{Multi-task graph classification.}
  \label{sfig:architecture-graph-classification}
\end{subfigure}
\caption{
    (a) The network architecture used for node classification on AIFB and
    MUTAG. This architecture is the same as in \cite{Schlichtkrull2018} except
    with \glspl{rgcn}
    replaced with \glspl{rgat}.
    (b) The network architecture used for multi-task graph classification on
    Tox21. This architecture is the same as the \glspl{gcn} architecture in
    \cite{Altae-Tran2016} except with \glspl{rgcn} replaces with \glspl{gat} and
    we do not use graph pooling.}
\label{fig:results}
\end{figure}

For the transductive task of semi-supervised node classification, we employ a
two-layer \gls{rgat} architecture shown in
\Cref{sfig:architecture-node-classification}. We use a \gls{relu} activation
after the \gls{rgat} concat layer, and a node-wise softmax on the final layer to
produce an estimate for the probability that the \ith \, label is in the class $\alpha$
\begin{equation}
  \label{eq:node-softmax}
  P(\text{class}_i=\alpha)
  \approx \hat y_{i,\alpha}
  = \softmax(\bmh_i^{(2)})_{\alpha}.
\end{equation}
We then employ a masked cross-entropy loss $\mcL$ to constrain the
network updates to the subset of nodes $\mcY$ whose labels are known
\begin{equation}
  \mcL=-\sum_{i\in\mcY}\sum_{\alpha=1}^{n_{\text{classes}}} y_{i,\alpha}\ln\left(\hat y_{i,\alpha}\right),
\end{equation}
where $\bmy_{i}$ is the one-hot representation of the true label for node $i$.

\subsection{Graph classification}
\label{sec:graph-classification}

For inductive graph classification, we employ a
two-layer \gls{rgat} followed by a graph gather and dense network architecture shown in
\Cref{sfig:architecture-graph-classification}. We use \gls{relu} activations
after each \gls{rgat} layer and the first dense layer. We use a $\tanh$
activation after the $\text{GraphGather}: \mbbR^{N\times F}\rightarrow \mbbR^{2F}$, which is a vector concatenation of the mean of the node representations with the
feature-wise $\max$ of the node representations
\begin{equation}
  \bmH^\prime
  = \text{GraphGather}(\bmH)
  = \left(\frac1N\sum_{i=1}^N\bmh_i\right)\oplus\left[ \bigoplus_{f=1}^F \max_i h_{i,f} \right].
\end{equation}
The final dense layer then produces logits of the size
$n_{\text{classes}}\times n_{\text{tasks}}$, and we apply a task-wise softmax to
its output to produce an estimate $\hat y_{t,\alpha}$ for the probability that
the graph is in class $\alpha$ for a given task $t$, analogous to
\Cref{eq:node-softmax}. Weighted cross-entropy loss
$\mcL$ is then used to form the learning objective
\begin{equation}
  \mcL(w, y, \hat y)=-\sum_{t=1}^{n_\text{tasks}}\sum_{\alpha=1}^{n_{\text{classes}}} w_{t,\alpha}\,y_{t,\alpha}\ln\left(\hat y_{t,\alpha}\right),
\end{equation}
where $w_{t,\alpha}$ and $y_{t,\alpha}$ are the weights and one-hot true labels for
task $t$ and class $\alpha$ respectively.

\section{Evaluation}
\label{sec:evaluation}
\subsection{Datasets}
\label{subsec:experiments/graphs}
We evaluate the models on transductive and inductive 
tasks. Following the experimental setup of \cite{Schlichtkrull2018} for the
transductive tasks, we evaluate our model on the \gls{rdf} datasets AIFB and
MUTAG. We also evaluate our model for an inductive task on the molecular
dataset, Tox21. Details of these data sets are given in Table
\ref{table:datasets}. For further details on the transductive and inductive datasets, please see \cite{Ristoski2016} and \cite{Wu2018} respectively.

\begin{table}[t]
\caption{A summary of the datasets used in our experiments and how they are partitioned.}
\centering
\begin{tabular}{@{}llll@{}}
\toprule
Datasets         & AIFB            & MUTAG            & Tox21                              \\ \midrule
Task             & Transductive    & Transductive     & Inductive                          \\
Nodes            & 8,285 (1 graph) & 23,644 (1 graph) & 145,459 (8014 graphs)              \\
Edges            & 29,043          & 74,227           & 151,095                            \\
Relations        & 45              & 23               & 4                                  \\
Labelled         & 176             & 340              & 96,168 (12 per graph)              \\
Classes          & 4               & 2                & 12 (multi-label)                   \\
Train nodes      & 112             & 218              & (6411 graphs)                      \\
Validation nodes & 28              & 54               & (801 graphs)                       \\
Test nodes       & 28              & 54               & (802 graphs)                       \\ \bottomrule
\end{tabular}
\label{table:datasets}
\end{table}

%
%
%
%

\paragraph{Transductive baselines}
We consider as a baseline the recent state-of-the-art results from
\cite{Schlichtkrull2018} obtained with a two-layer RGCN model with 16 hidden
units and basis function decomposition. We also include the same challenging
baselines of FEAT \citep{Paulheim2012}, WL \citep{Shervashidze2011,deVries2015}
and RDF2Vec \citep{Ristoski2016}. In-depth details of these baselines are given
by \cite{Ristoski2016}.

\paragraph{Inductive baselines} As baselines for Tox21, we
compare against the most competitive methods on Tox21 reported in \cite{Wu2018}.
Specifically, we compare against deep multitask networks \cite{Ramsundar2015}, deep bypass multitask networks \cite{Wu2018}, Weave \cite{Kearnes2016}, and a
\gls{rgcn} model whose relational structure is determined by the degree of the
node to be updated \cite{Altae-Tran2016}. Specifically, up to and including some maximum
degree $D_{\text{max}}$,
\begin{equation}
  \bmh^\prime_i
  =\sigma\left[
    (\bmW^{\text{deg(i)}})^T \bmh_i +
    \sum_{j\in\mcN_i}(\bmU^{\text{deg{(i)}}})^T\bmh_j + \bmb^{\text{deg}(i)}\right],
\end{equation}
where $\bmW^{\text{deg}(i)}\in\mbbR^{F\times F^\prime}$ is a degree-specific
linear transformation for self-connections,
$\bmU^{\text{deg(i)}}\in\mbbR^{F\times F^\prime}$ is a
degree-specific linear transformation for neighbours into their intermediate
representations $\bmg_i\in\mbbR^{F^\prime}$, and $\bmb^{\text{deg}(i)}$ is a
degree-specific bias. Any update for any degree $d(i)>D_{\text{max}}$ gets assigned to the
update for the maximum degree $D_{\text{max}}$.

\subsection{Experimental setup}
\label{subsec:experiments/setup}

\paragraph{Transductive learning} For the transductive learning tasks, 
the architecture discussed in \Cref{sec:node-classification} was applied. Its
hyperparameters were optimised for both AIFB and MUTAG on their respective
training/validation sets defined in \cite{Ristoski2016}, using 5-fold cross
validation. Using the found hyperparameters, we retrain on the full training set
and report results on the test set across 200 seeds. We employ early stopping on
the validation set during cross-validation to determine the number of epochs we
will run on the final training set. Hyperparameter optimisation details are
given in \Cref{tab:hyperparameters-transductive} of \Cref{app:hyperparameters}.

\paragraph{Inductive learning} For the inductive learning tasks,
the architecture discussed in \Cref{sec:graph-classification} was applied. In order to optimise hyperparameters once, ensure no data leakage, but also provide
comparable benchmarks to those presented in \cite{Wu2018}, three benchmark
splits were taken from the \texttt{MolNet benchmarks}\footnote{Retrieved from
\url{http://deepchem.io.s3-website-us-west-1.amazonaws.com/trained_models/Hyperparameter_MoleculeNetv3.tar.gz}.},
and graphs belonging to any of the test sets were isolated. Using the remaining graphs we
performed a hyperparameter search using 2 folds of 10-fold cross validation. Using
the found hyperparameters, we then retrained on the three benchmark splits
provided with 2 seeds each, giving an unbiased estimate of model performance. We
employ early stopping during both the cross-validation and final run (the
validation set of the inductive task is available for the final benchmark, in
contrast to the transductive tasks) to determine the number of training epochs.
Hyperparameter optimisation details are
given in \Cref{tab:hyperparameters-inductive} of \Cref{app:hyperparameters}.

\paragraph{Constant attention} In all experiments, we train with the attention mechanism turned on. At evaluation
time, however, we report results with and without the attention mechanism to
provide insight into whether the attention mechanism helps. \gls{argat} (\gls{wirgat})
without the attention is called C-\gls{argat} (C-\gls{wirgat}).
\subsection{Results}
\label{subsec:evaluation/results}
\begin{table}[t]
  \caption{(a) Entity classification results accuracy (mean and standard deviation
    over 10 seeds) for
    FEAT \citep{Paulheim2012},
    WL \citep{Shervashidze2011,deVries2015},
    RDF2Vec \citep{Ristoski2016} and
    \gls{rgcn} \citep{Schlichtkrull2018},
    and (mean and standard deviation over
    200 seeds) for our implementation of \gls{rgcn}, as well as additive and
    multiplicative attention for (C-)\gls{wirgat} and (C-)\gls{argat}
  (this work). Test performance is
  reported on the splits provided in
  \cite{Ristoski2016}. (b) Graph classification mean \gls{roc} \gls{auc} across all 12
  tasks (mean and standard deviation
  over 3 splits) for
  Multitask \citep{Ramsundar2015},
  Bypass \citep{Wu2018},
  Weave \citep{Kearnes2016},
  \gls{rgcn} \citep{Altae-Tran2016},
  and (mean and standard deviation
  over 3 splits, 2 seeds per split) our implementation of \gls{rgcn},
  additive and multiplicative attention for
  (C-)\gls{wirgat} and (C-)\gls{argat} (this work). Test
  performance is reported on the
  splits provided in \cite{Wu2018}. Best performance in class in boldened, and best performance overall is underlined. For completeness, we present the training
  and validation mean \gls{roc}-\glspl{auc} alongside the test
  \gls{roc}-\glspl{auc} in \Cref{app:tox21-results}.
  For a graphical representation of these results, see \Cref{fig:charts} in
  \Cref{app:charts}.}
  \begin{subfigure}{.5\textwidth}
    \centering
    \caption{Transductive}
    \begin{tabular}{lcccc}
      \toprule
      Model        & AIFB                      & MUTAG                    \\ \midrule
      Feat         & $55.55\pm0.00$            & $77.94 \pm 0.00$         \\
      WL           & $80.55\pm0.00$            & $\underline{\mathbf{80.88}}\pm 0.00$ \\
      RDF2Vec      & $88.88 \pm 0.00$          & $67.20 \pm 1.24$         \\
      RGCN        & $\mathbf{95.83} \pm 0.62$ & $73.23 \pm 0.48$ \\ \midrule
      RGCN (ours) & $94.64 \pm 2.75$ & $74.15 \pm 2.40$ \\ \midrule \midrule
      \multicolumn{3}{l}{\emph{Additive attention}} \\ \midrule
      C-\gls{wirgat} & $\underline{\mathbf{96.86}} \pm 0.94$ & $69.37 \pm 2.75$ \\
      \gls{wirgat}   & $96.83 \pm 1.01$ & $\mathbf{69.83} \pm 2.74$ \\
      C-\gls{argat}  & $93.05 \pm 3.05$ & $63.69 \pm 8.41$ \\
      \gls{argat}    & $94.01 \pm 2.76$ & $65.54 \pm 6.25$ \\ \midrule \midrule
      \multicolumn{3}{l}{\emph{Multiplicative attention}} \\ \midrule
      C-\gls{wirgat} & $93.71 \pm 3.33$ & $69.57 \pm 3.70$ \\
      \gls{wirgat}   & $92.92 \pm 3.75$ & $69.60 \pm 3.75$ \\
      C-\gls{argat}  & $95.89 \pm 1.93$ & $\mathbf{74.38} \pm 3.78$ \\
      \gls{argat}    & $\underline{\mathbf{96.19}} \pm 1.70$ & $73.17 \pm 3.41$ \\ 
      \bottomrule
    \end{tabular}
    \label{subtable:transductive}
  \end{subfigure} 
  \begin{subfigure}{.5\textwidth}
    \centering
    \caption{Inductive}
    \begin{tabular}{lc}
      \toprule
      Model              & Tox21 \\ \midrule
      Multitask          & $0.803\pm0.012$ \\
      Bypass             & $0.810\pm0.013$ \\
      Weave              & $0.820\pm0.010$ \\
      RGCN               & $0.829\pm0.006$ \\ \midrule
      RGCN (ours)        & $\mathbf{0.835}\pm0.008$ \\ \midrule \midrule
      \multicolumn{2}{l}{\emph{Additive attention}} \\ \midrule
      C-\gls{wirgat}     & $0.832 \pm 0.009$ \\
      \gls{wirgat}       & $\mathbf{0.835} \pm 0.006$ \\
      C-\gls{argat}      & $0.829 \pm 0.010$ \\
      \gls{argat}        & $\mathbf{0.835} \pm 0.006$ \\ \midrule \midrule
      \multicolumn{2}{l}{\emph{Multiplicative attention}} \\ \midrule
      C-\gls{wirgat}     & $0.811 \pm 0.008$ \\
      \gls{wirgat}       & $\underline{\textbf{0.838}} \pm 0.007$ \\
      C-\gls{argat}      & $0.802 \pm 0.007$ \\
      \gls{argat}        & $0.837 \pm 0.007$ \\
      \bottomrule
    \end{tabular}  
    \label{subtable:inductive}
  \end{subfigure}
  \label{tab:all-results}
\end{table}

\subsubsection{Benchmarks and additional analyses}

Model means and standard deviations are presented in \Cref{tab:all-results}. To
provide a picture of characteristic model behaviour, the \glspl{cdf} for
the hyperparameter sweep are presented in \Cref{fig:all-cdfs} of
\Cref{app:cdfs}. To draw meaningful conclusions, we compare against
our own implementation of \gls{rgcn} rather than the results reported in
\cite{Schlichtkrull2018,Wu2018}.

We will occasionally employ a one-sided hypothesis test in order to make
concrete statements about model performance.
The details and complete results of this test are presented in
\Cref{app:significance}. When we refer to significant results this corresponds
to a test statistic supporting our hypothesis with a $p-$value $p\leq0.05$.

\subsubsection{Transductive learning}

In \Cref{subtable:transductive} we evaluate
\gls{rgat} on MUTAG and AIFB.
With additive attention, \gls{wirgat} outperforms \gls{argat}, consistent with
\cite{Schlichtkrull2018}. Interestingly, when employing multiplicative
attention, the converse appears true. For node classification tasks on \gls{rdf} data,
this indicates that the importance of a particular relation type does not vary
much (if at all) across the graph unless one employs a multiplicative
comparison\footnote{Or potentially other comparisons beyond additive or
  constant, i.e. \gls{rgcn}.}
between node representations.

\paragraph{AIFB} On AIFB, the best to worst performing models are: 1) additive
\gls{wirgat} 2) multiplicative \gls{argat}
3) \gls{rgcn} 4) additive \gls{argat}, and 5) multiplicative \gls{wirgat}, with each
comparison being significant.

When comparing against their constant attention counterparts, the
significant differences observed were for additive and multiplicative
\gls{argat}, where attention gives a relative mean performance improvements of
1.03\% and 0.31\% respectively, and multiplicative \gls{wirgat}, where attention
gives a relative mean performance drop of 0.84\%.

Although we present state-of-the art result on AIFB with additive \gls{wirgat},
since its performance with and without attention are not significantly
different, it is unlikely that this is due to the attention mechanism itself, at
least at inference time. Over the hyperparameter space, additive \gls{wirgat}
and \gls{rgcn} are comparable in performance (see \Cref{subfig:aifb-cdf} in
\Cref{app:cdfs}), leading us to conclude that the result is more likely
attributable to finding a better hyperparameter point for additive \gls{wirgat}
during the search.

\paragraph{MUTAG} On MUTAG, the best to worst performing models are: 1)
\gls{rgcn} 2) multiplicative \gls{argat} 3) additive \gls{wirgat} tied with
multiplicative \gls{wirgat}, and 4) additive \gls{argat}, with each comparison
being significant.

When comparing against their constant attention counterparts, the
significant differences observed were for additive \gls{wirgat} and \gls{argat}, where attention gives relative mean performance improvements of
0.66\% and 2.90\% respectively, and multiplicative \gls{argat}, where attention
gives a relative mean performance drop of 1.63\%.

We note that \gls{rgcn} consistently outperforms \gls{rgat} on MUTAG, contrary
to what might be expected \citep{Schlichtkrull2018}. The result is
surprising given that \gls{rgcn} lies within the parameter space of
\gls{rgat} (where the attention kernel is zero), a configuration we check
through evaluating C-\gls{wirgat}. In our
experiments we have observed that both \gls{rgcn} and \gls{rgat} can memorise
the MUTAG training set with $100\%$ accuracy without difficulty (this is
not the case for AIFB). The performance gap between \gls{rgcn} and \gls{rgat}
could then be explained by the following:
\begin{itemize}
  \item[-] During training, the \gls{rgat} layer uses its attention
    mechanism to solve the learning objective. Once the objective is solved, the
    model is not encouraged by the loss function to seek a point in the parameter
    space that would also behave well when attention is set to a normalising
    constant within neighbourhoods (i.e. the parameter space point that would be
    found by \gls{rgcn}).
  \item[-] The \gls{rdf} tasks are transductive, meaning that a basis-dependent
    spectral approach is sufficient to solve them. As \gls{rgcn} already
    memorises the MUTAG training set, a model more
    complex\footnote{Measured in terms of \gls{mdl}, for example.} than
    \gls{rgcn}, for example \gls{rgat}, that can also
    memorise the training set is unlikely to generalise as well, although this
    is a hotly debated topic - see e.g. \cite{Zhang2016}.
\end{itemize}
We employed a suite of regularisation techniques to get \gls{rgat} to generalise
on MUTAG, including L2-norm penalties, dropout in multiple places, batch
normalisation, parameter reduction and early stopping, however, no evaluated harshly
regularised points for \gls{rgat} generalise well on MUTAG.

Our final observation is that the attention mechanism presented
in \Cref{subsec:rgat-architecture/relational-graph-attention-layer} relies on
node features. The node features for the above tasks are learned from
scratch (the input feature matrix is a one-hot node index) as
part of the task. It is possible that in this semi-supervised setup,
there is insufficient signal in the data to learn both the input node
embeddings as well as a meaningful attention mechanism to act upon them.

\subsubsection{Inductive learning}

In \Cref{subtable:inductive} we evaluate
\gls{rgat} on Tox21. The number of samples is lower for these evaluations
than for the transductive tasks, and so fewer model comparisons will be
accompanied with a reasonable significance, although there are still some
conclusions we can draw.

Through a thorough hyperparameter search, and incorporating various
regularisation techniques, we obtained the relative mean performance of
0.72\% for \gls{rgcn} compared to the result reported in \cite{Wu2018},
providing a much stronger baseline.

Both additive attention models match the performance of \gls{rgcn}, whereas
multiplicative \gls{wirgat} and \gls{argat} marginally outperform \gls{rgcn}, although this
is not significant ($p=0.24$ and $p=0.41$ respectively).

When comparing against their constant attention counterparts,
significant differences observed were for multiplicative \gls{wirgat} and
\gls{argat},
where attention gives a relative mean performance improvements of
3.33\% and 4.36\% respectively. We do not observe any significant gains coming
from additive attention when compared to their constant counterparts. 

\section{Conclusion}
\label{sec:conclusion}
\glsresetall
We have investigated a class of models we call
\glspl{rgat}. These models act upon graph structures,
inducing a masked self-attention that takes account of local relational
structure as well as node features. This allows both nodes and their
properties under specific relations to be dynamically assigned an
importance for different nodes in the graph, and opens up graph attention
mechanisms to a wider variety of problems.

We evaluted two specific attention mechanisms, \gls{wirgat} and \gls{argat},
under both an additive and multiplicative logit construction, and compared them
to their equivalently evaluated spectral counterpart \glspl{rgcn}.

We find \glspl{rgat} perform competitively or poorly on established
baselines. This behavior appears
strongly task-dependent. Specifically, relational inductive tasks such as graph
classification benefit from multiplicative \gls{argat}, whereas transductive
relational tasks, such as knowledge base completion, at least in the
absence of node features, are better tackled using spectral methods like
\glspl{rgcn} or other graph feature extraction methods like \gls{wl} graph
kernels.

In general we have found that \gls{wirgat} should be paired with an additive
logit mechanism, and fares marginally better than \gls{argat} on transductive
tasks, whereas \gls{argat} should be paired with a multiplicative logit
mechanism, and fares marginally better on inductive tasks.

We have found no cases where choosing any variation of
\gls{rgat} is guaranteed to significantly outperform \gls{rgcn}, although we
have found that in cases where \gls{rgcn} can memorise the training set, we are
confident that \gls{rgat} will not perform as well
as \gls{rgcn}. Consequently, we suggest that before attempting to train
\gls{rgat}, a good first test is to inspect the training set performance of
\gls{rgcn}.

Through our thorough evaluation and presentation of the behaviours and
limitations of these models, insights can be derived that will enable the
discovery of more powerful model architectures that act upon relational
structures.
Observing that model variance on all of the tasks presented here is high, any
future work developing and expanding these methods must choose larger, more
challenging datasets. In addition, a comparison between the generalisation of
spectral methods, like those presented here, and generalisations of \glspl{rnn},
like Gated Graph Sequence Networks, is a necessary ingredient for determining the
most promising future direction for these models.

\section{Acknowledgements}
\label{sec:acknowledgements}
We thank Ozan Oktay for many fruitful discussions during the early stages of this
work, and Jeremie Vallee for assistance with the experimental setup. We also
thank April Shen, Kostis Gourgoulias and Kristian Boda, whose comments greatly improved
the manuscript, as well as Claire Woodcock for support.

\bibliographystyle{humannat}

\newpage
\appendix
\section{Tox21 Results}
\label{app:tox21-results}

For completeness, we present the training, validation and test set performance
of our models in addition to those in 
\cite{Wu2018} in \Cref{table:tox21-results}.
\begin{table}[ht]
\caption{Graph classification mean \gls{auc} across all 12
  tasks (mean and standard deviation
  over 3 splits) for
  Multitask \citep{Ramsundar2015},
  Bypass \citep{Wu2018},
  Weave \citep{Kearnes2016},
  \gls{rgcn} \citep{Altae-Tran2016}, our implementation of \gls{rgcn}, additive
  and multiplicative attention versions of 
  \gls{wirgat} and \gls{argat} (this work). Training, validation and test
  performance is reported on the
  splits provided in \cite{Wu2018}. Best performance in class in boldened, and best performance overall is underlined.}
\centering
\vspace{0.1cm}
\begin{tabular}{lcccc}
 \toprule
  Model            & Training        & Validation                        & Test            \\ \midrule
  Multitask        & $0.884\pm0.001$ & $0.795\pm0.017$                   & $0.803\pm0.012$ \\
  Bypass           & $0.938\pm0.001$ & $0.800\pm0.008$                   & $0.810\pm0.013$ \\
  Weave            & $0.875\pm0.004$ & $0.828\pm0.008$                   & $0.820\pm0.010$ \\
  RGCN             & $\underline{\mathbf{0.905}}\pm0.004$ & $0.825\pm0.013$          & $0.829\pm0.006$ \\ \midrule
  RGCN (ours)      & $0.883\pm0.010$ & $\mathbf{0.845}\pm0.003$   & $\mathbf{0.835}\pm0.008$ \\ \midrule \midrule
  \multicolumn{2}{l}{\emph{Additive attention}} \\ \midrule
  C-\gls{wirgat}   & $0.897\pm0.022$ & $0.842\pm0.004$ & $0.832\pm0.009$ \\
  \gls{wirgat}     & $\mathbf{0.902}\pm0.024$ & $0.845\pm0.005$ & $\mathbf{0.835}\pm0.006$ \\
  C-\gls{argat}    & $0.884\pm0.012$ & $0.848\pm0.003$ & $0.829\pm0.010$ \\
  \gls{argat}      & $0.896\pm0.016$ & $\mathbf{0.851}\pm0.004$ & $\mathbf{0.835}\pm0.006$ \\ \midrule \midrule
  \multicolumn{2}{l}{\emph{Multiplicative attention}} \\ \midrule 
  C-\gls{wirgat}   & $0.859\pm0.016$ & $0.830\pm0.007$ & $0.811\pm0.008$ \\
  \gls{wirgat}     & $\mathbf{0.904}\pm0.022$ & $\underline{\mathbf{0.852}}\pm0.002$ & $\underline{\textbf{0.838}}\pm0.007$ \\
  C-\gls{argat}    & $0.838\pm0.007$ & $0.816\pm0.007$ & $0.802\pm0.007$ \\
  \gls{argat}      & $0.802\pm0.007$ & $0.846\pm0.003$ & $0.837\pm0.007$ \\ \bottomrule
\end{tabular}
\label{table:tox21-results}
\end{table}

\newpage
\section{Hyperparameters}
\label{app:hyperparameters}

We perform hyperparameter optimisation using \texttt{hyperopt}
\cite{pmlr-v28-bergstra13} with priors for the transductive tasks specified in
\Cref{tab:hyperparameters-transductive} and priors for the inductive tasks
specified in \Cref{tab:hyperparameters-inductive}.
In all experiments we use the Adam optimiser \citep{Kingma2014}.

\begin{table}[ht]
\caption{Priors on the hyperparameter search space for the transductive tasks.
  When multihead attention is used, the number of units per head is
  appropriately reduced in order to keep the total number of output units of an
  \gls{rgat} layer independent of the number of heads.}
\centering
\vspace{0.1cm}
\begin{tabular}{ll}
 \toprule
  Hyperparameter       & Prior \\ \midrule
  Graph kernel units   & $\text{MultiplesOfFour}(4, 20)$ \\
  Heads                & $\text{OneOf}(1,2,4)$ \\
  Feature dropout rate & $\text{Uniform}(0.0, 0.8)$            \\
  Edge dropout         & $\text{Uniform}(0.0, 0.8)$            \\ 
  $\bmW$ basis size    & $\text{OneOf}(\text{Full},5,10,20,30)$ \\
  Graph layer 1 $\bmW$ L2 coef   & $\text{LogUniform}(10^{-6}, 10^{-1})$ \\
  Graph layer 2 $\bmW$ L2 coef   & $\text{LogUniform}(10^{-6}, 10^{-1})$ \\
  $\bmA$ basis size    & $\text{OneOf}(\text{Full},5,10,20,30)$ \\
  Graph layer 1 $\bmA$ L2 coef   & $\text{LogUniform}(10^{-6}, 10^{-1})$  \\
  Graph layer 2 $\bmA$ L2 coef   & $\text{LogUniform}(10^{-6}, 10^{-1})$ \\
  Learning rate        & $\text{LogUniform}(10^{-5}, 10^{-1})$\\
  Use bias & $\text{OneOf}(\text{Yes}, \text{No})$\\
  Use batch normalisation & $\text{OneOf}(\text{Yes}, \text{No})$\\
\bottomrule
\end{tabular}
\label{tab:hyperparameters-transductive}
\end{table}

\begin{table}[ht]
\caption{Priors on the hyperparameter for the inductive
  task. The batch size was held at 64, and no bases decomposition is used.
  When multihead attention is  used, the number of units per head is
  appropriately reduced in order to keep
  the total number of output units of an \gls{rgat} layer independent of the
  number of heads.}

\centering
\vspace{0.1cm}
\begin{tabular}{ll}
 \toprule
  Hyperparameter      & Prior \\ \midrule
  Graph kernel units  & $\text{MultiplesOfEight}(32, 128)$ \\
  Dense units         & $\text{MultiplesOfEight}(32, 128)$ \\
  Heads               & $\text{OneOf}(1,2,4,8)$\\
  Feature dropout     & $\text{Uniform}(0.0, 0.8)$            \\
  Edge dropout        & $\text{Uniform}(0.0, 0.8)$            \\ 
  $\bmW$ L2 coef (1)  & $\text{LogUniform}(10^{-6}, 10^{-1})$   \\
  $\bmW$ L2 coef (2)  & $\text{LogUniform}(10^{-6}, 10^{-1})$   \\
  $\bma$ L2 coef (1)  &  $\text{LogUniform}(10^{-6}, 10^{-1})$  \\
  $\bma$ L2 coef (2)  &   $\text{LogUniform}(10^{-6}, 10^{-1})$ \\
  Learning rate & $\text{LogUniform}(10^{-5}, 10^{-1})$\\
  Use bias & $\text{OneOf}(\text{Yes}, \text{No})$\\
  Use batch normalisation & $\text{OneOf}(\text{Yes}, \text{No})$\\
\bottomrule
\end{tabular}
\label{tab:hyperparameters-inductive}
\end{table}

\newpage
\section{Charts}
\label{app:charts}

\begin{figure}[ht]
  \begin{subfigure}{.5\textwidth}
    \centering
      \includegraphics[width=1.1\textwidth]{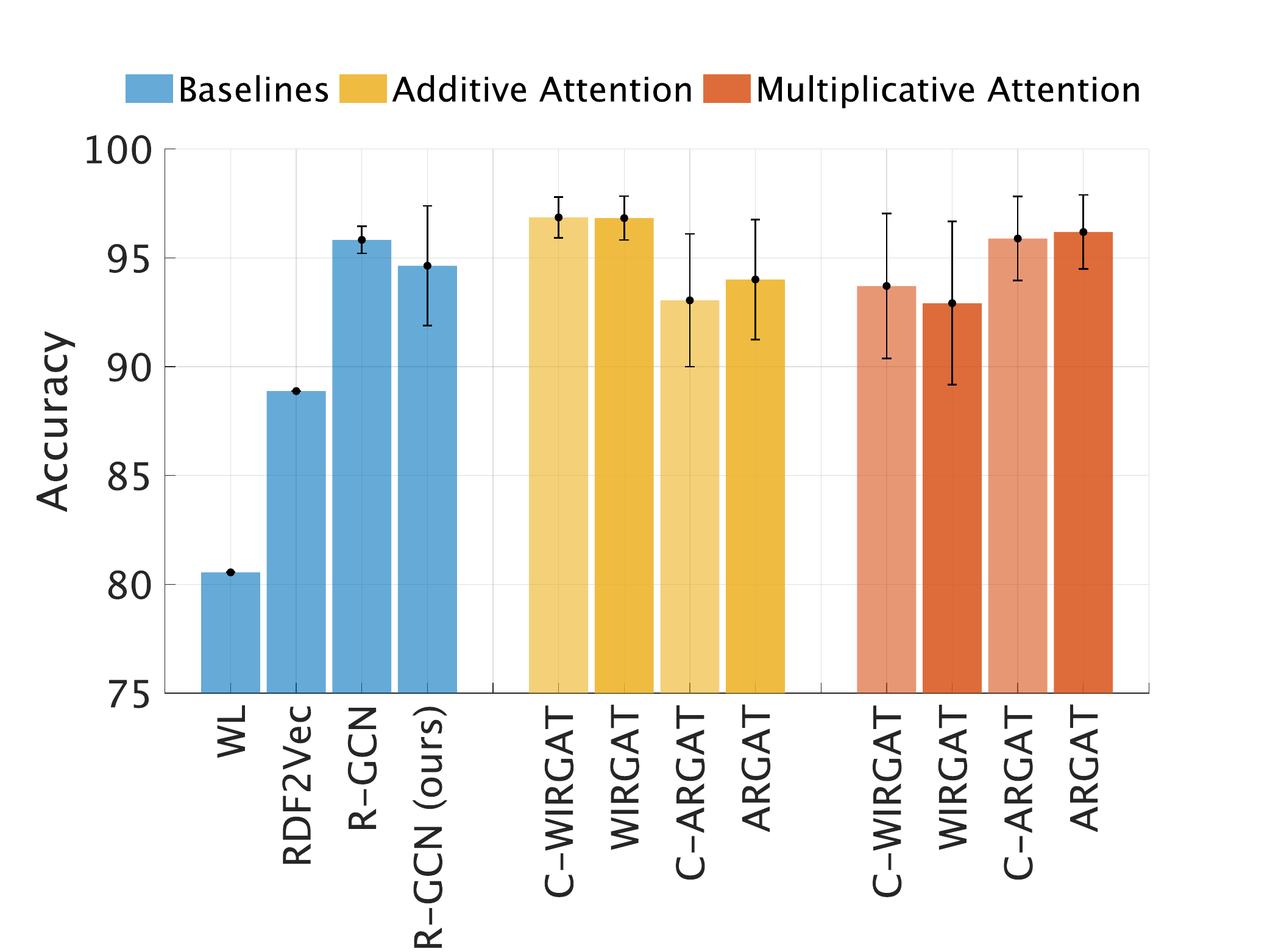}
    \caption{AIFB}
    \label{subfig:aifb}
  \end{subfigure} 
  \begin{subfigure}{.5\textwidth}
    \centering
      \includegraphics[width=1.1\textwidth]{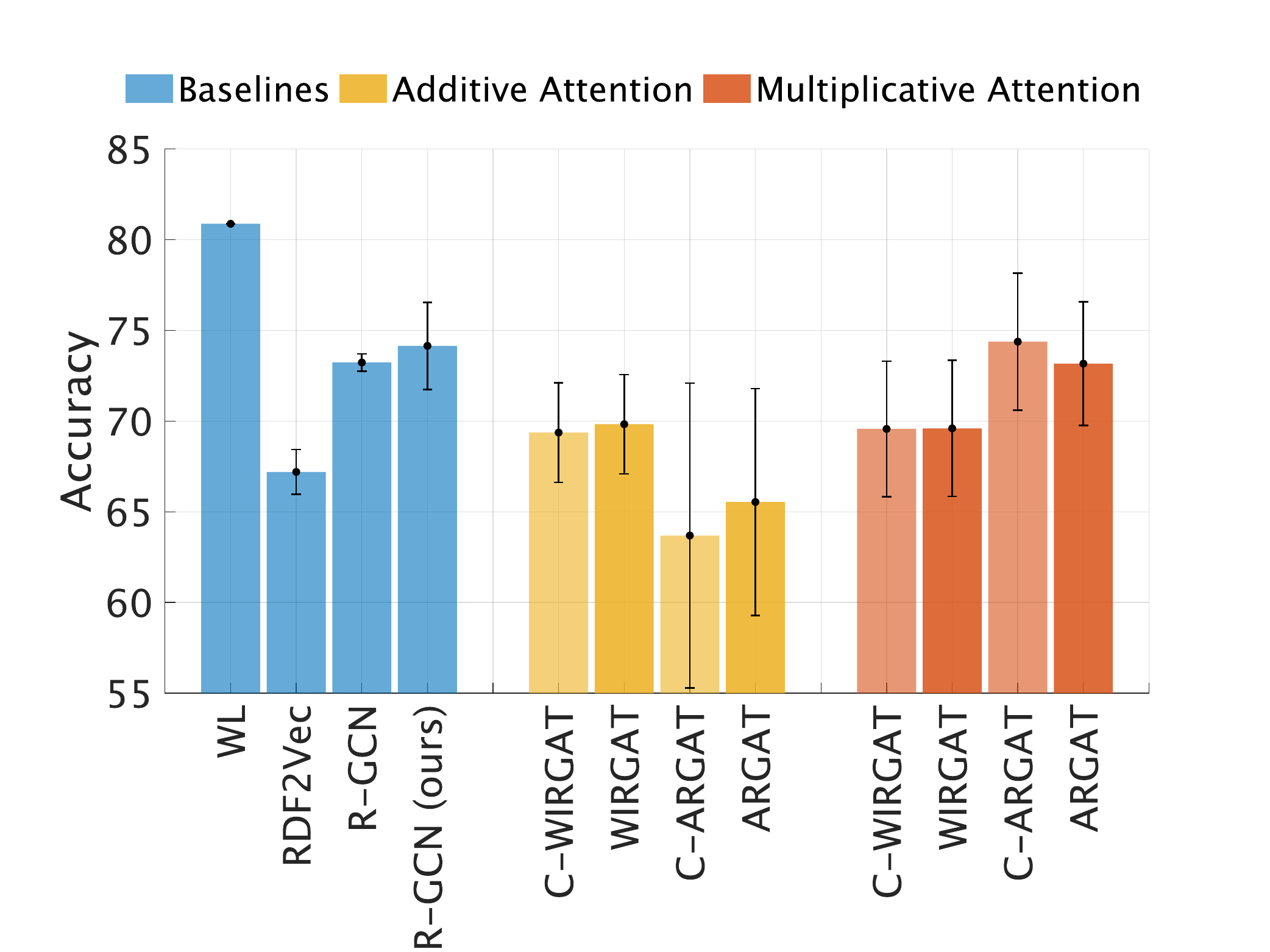}
    \caption{MUTAG}
    \label{subfig:mutag}
  \end{subfigure}\\
  \begin{center}
    \begin{subfigure}{.5\textwidth}
      \centering
        \includegraphics[width=1.1\textwidth]{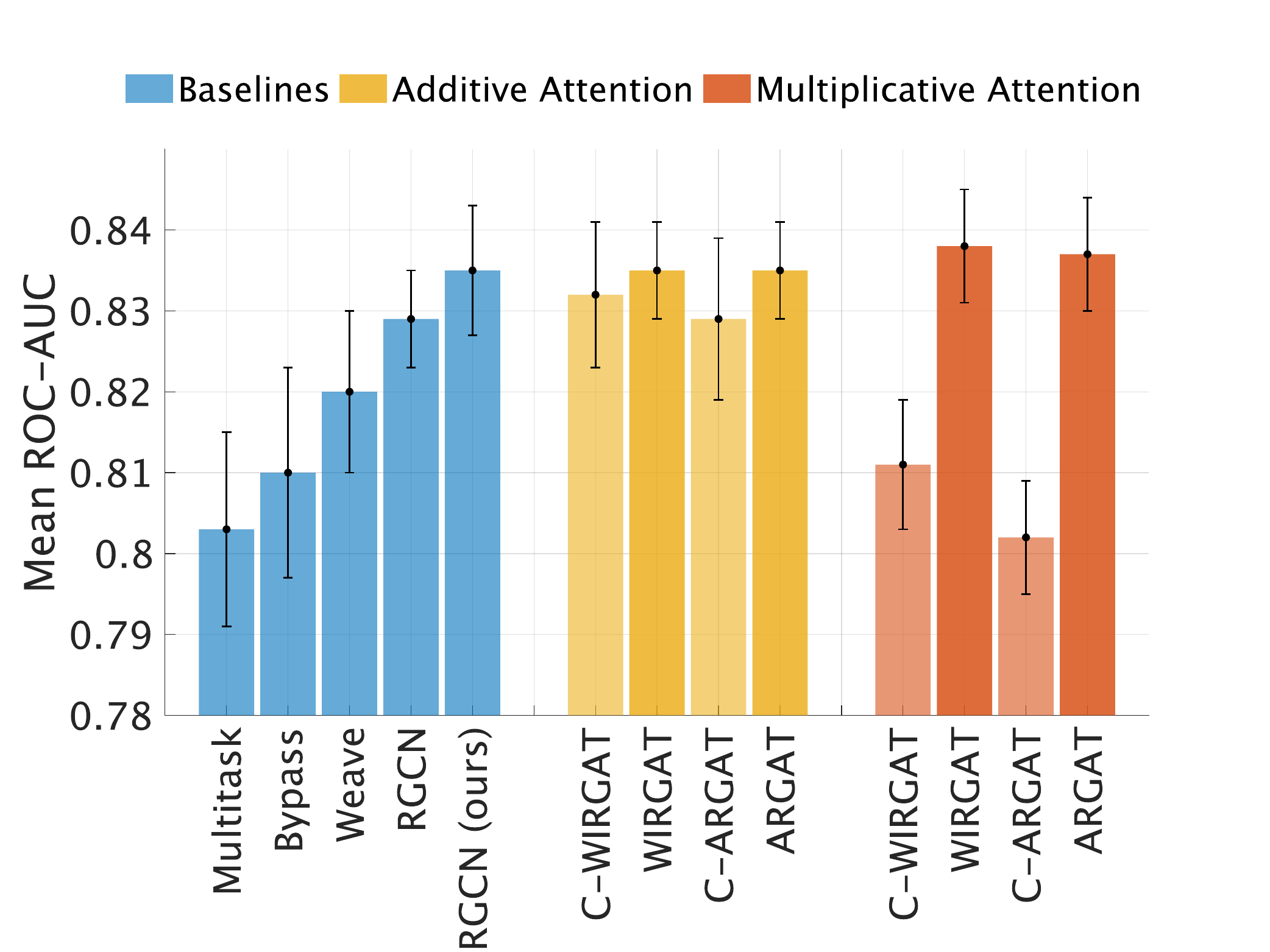}
      \caption{TOX21}
      \label{subfig:tox}
    \end{subfigure}
  \end{center}
  \caption{(a) and (b): \textbf{Blue}
    Baseline entity classification accuracy (mean and
    standard deviation over 10 seeds) for
    FEAT \citep{Paulheim2012},
    WL \citep{Shervashidze2011,deVries2015},
    RDF2Vec \citep{Ristoski2016} and
    \gls{rgcn} \citep{Schlichtkrull2018},
    and (mean and standard deviation over
    200 runs) for our implementation of \gls{rgcn}.
    \textbf{Yellow}
    Entity classification accuracy (mean and
    standard deviation over 200 seeds) for additive attention (this work).
    \textbf{Red}
    Entity classification accuracy (mean and
    standard deviation over 200 seeds) for multiplicative attention (this work).
    Test performance is reported on the splits provided in \cite{Ristoski2016}.
    (c): \textbf{Blue} Baseline graph classification mean \gls{roc} \gls{auc} across all 12
    tasks (mean and standard deviation
    over 3 splits) for
    Multitask \citep{Ramsundar2015},
    Bypass \citep{Wu2018},
    Weave \citep{Kearnes2016},
    \gls{rgcn} \citep{Altae-Tran2016},
    and (mean and standard deviation
    over 3 splits, 2 seeds per split) our implementation of \gls{rgcn}.
    \textbf{Yellow}
    Additive attention graph classification mean \gls{roc}-\gls{auc} (mean and
    standard deviation over 200 seeds) across all 12 tasks (this work).
    \textbf{Red}
    Multiplicative attention graph classification mean \gls{roc}-\gls{auc} (mean and
    standard deviation over 200 seeds) across all 12 tasks (this work).
    All raw values are given in \Cref{tab:all-results}.}
  \label{fig:charts}
\end{figure}

To aid interpretability of the results presented in \Cref{tab:all-results} we present a chart
representation in \Cref{fig:charts}.

\newpage
\section{Cumulative distribution functions}
\label{app:cdfs}

To aid further insight into our results, we present the \glspl{cdf} for each
model on each task in \Cref{fig:charts}. In this context, we treat the
performance metric of interest during the hyperparameter search as the empirical
distribution of some random variable $X$. We then define
its \gls{cdf} $F_X(x)$ in the standard way  
\begin{equation}
  F_X(x)
  =P(X\leq x),
\end{equation}
where $P(X\leq x)$ is the probability that $X$ takes on a value less than or
equal to $x$. The \gls{cdf} allows one to gauge whether any given architecture
typically performs better than another across the whole space, rather than
comparison of the tuned hyperparameters, which in some cases may be outliers
in terms of generic behavior for that architecture.

\begin{figure}[ht]
  \begin{subfigure}{.5\textwidth}
    \centering
      \includegraphics[width=1.1\textwidth]{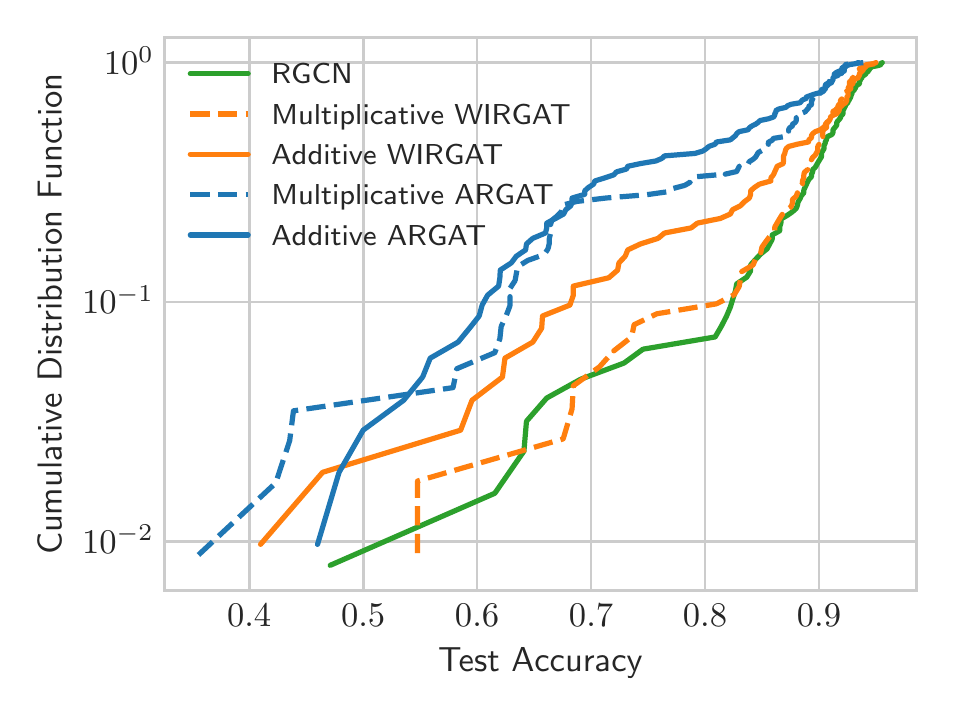}
    \caption{AIFB}
    \label{subfig:aifb-cdf}
  \end{subfigure} 
  \begin{subfigure}{.5\textwidth}
    \centering
      \includegraphics[width=1.1\textwidth]{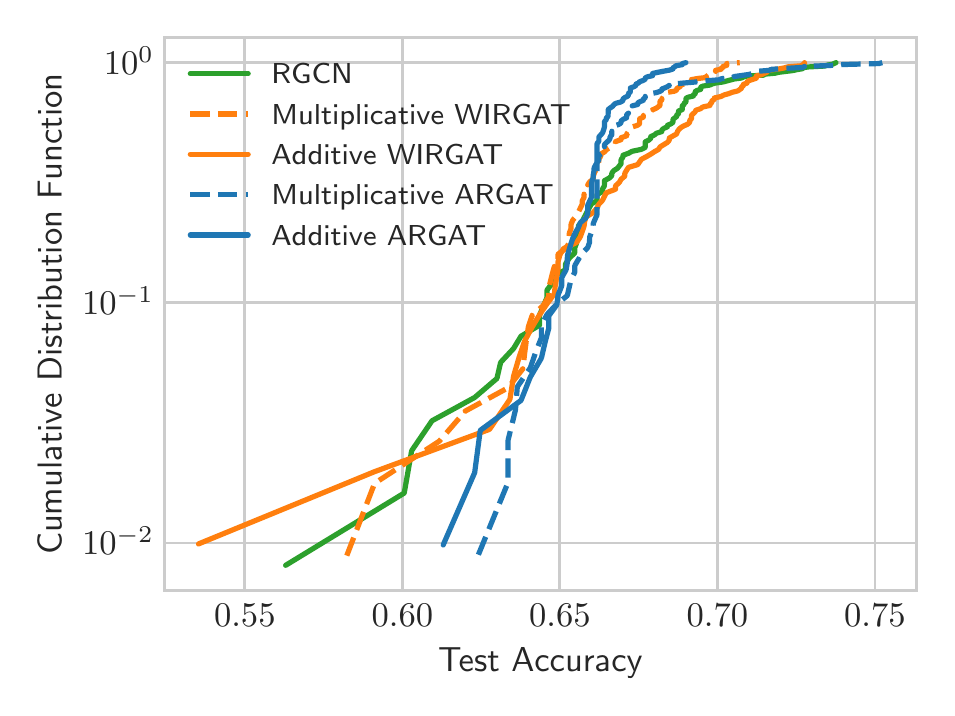}
    \caption{MUTAG}
    \label{subfig:mutag-cdf}
  \end{subfigure}\\
  \begin{center}
    \begin{subfigure}{.5\textwidth}
      \centering
        \includegraphics[width=1.1\textwidth]{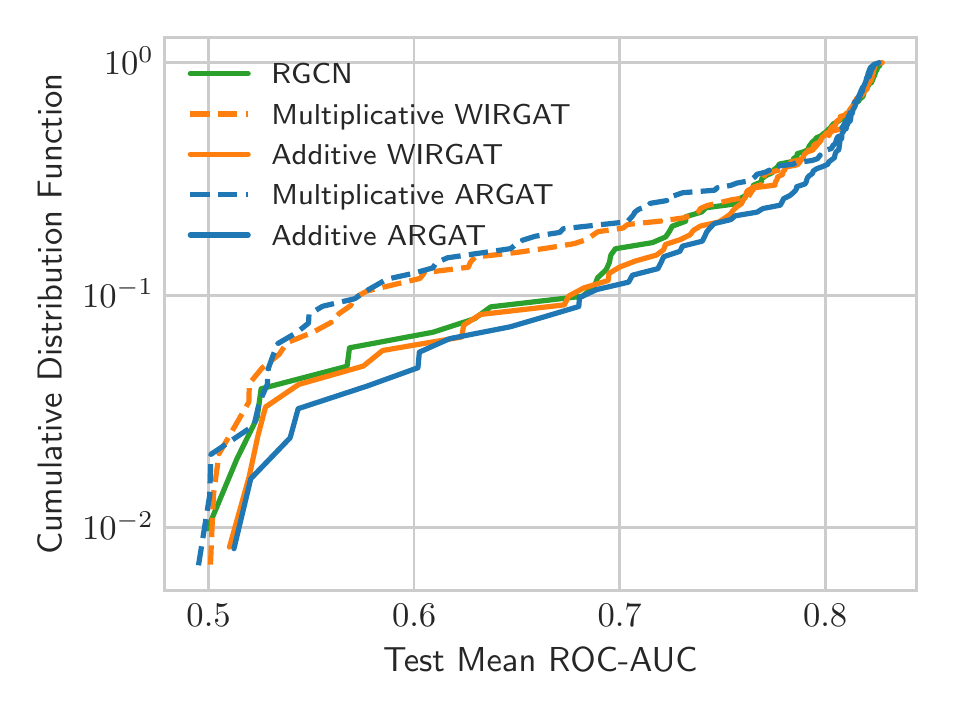}
      \caption{TOX21}
      \label{subfig:tox-cdf}
    \end{subfigure}
  \end{center}
  \caption{\glspl{cdf} for all models on a) AIFB, b) MUTAG and c) TOX21. Green
    lines correspond to our implementation of \gls{rgcn}, blue lines correspond
    to \gls{argat}, and orange lines correspond to \gls{wirgat}. Solid lines
    correspond to additive attention (and \gls{rgcn}), whereas dashed lines
    correspond to multiplicative attention. A lower \gls{cdf} value is better in the
    sense that a greater proportion of models of achieve a higher value of that
    metric.}
  \label{fig:all-cdfs}
\end{figure}

\paragraph{AIFB} Additive and multiplicative \gls{argat} perform poorly for most areas of the hyperparameter space, whereas \gls{rgcn} and multiplicative \gls{wirgat} perform comparably across the entire hyperparameter space.

\paragraph{MUTAG} Interestingly, the models that have a greater amount of hyperparameter space covering poor performance (i.e. \gls{rgcn}, multiplicative and additive \gls{wirgat}) are also the models which also have a greater amount of hyperparameter space covering good performance. In other words, on the MUTAG, the \gls{argat} prior resulted in a model whose test set performance was relatively insensitive to hyperparameter choice when compared against the other candidates. Given that the \gls{argat} model was the most flexible of the models evaluated, and that it was able to memorise the training set, this suggests that the task contained insufficient information for the model to learn its attention mechanism. Given that \gls{wirgat} was able to at least partially learn to its attention mechanism suggests that \gls{wirgat} is less data hungry than \gls{argat}.

\paragraph{Tox21} The multiplicative attention models fare poorly on the majority of the hyperparameter space compared to the other models. There is a slice of the hyperparameter space where the multiplicative attention models outperform the other models, however, indicating that although they are difficult to train, it may be worth spending time hyperoptimising them if you need the best performing model on a relational inductive task. The additive attention models and \gls{rgcn} perform comparably across the entirety of the hyperparameter space and generally perform better than the multiplicative methods except for the very small region of hyperparameter space mentioned above.

\newpage
\section{Significance testing}
\label{app:significance}
\begin{figure}[ht]
  \centering
    \includegraphics[width=\textwidth]{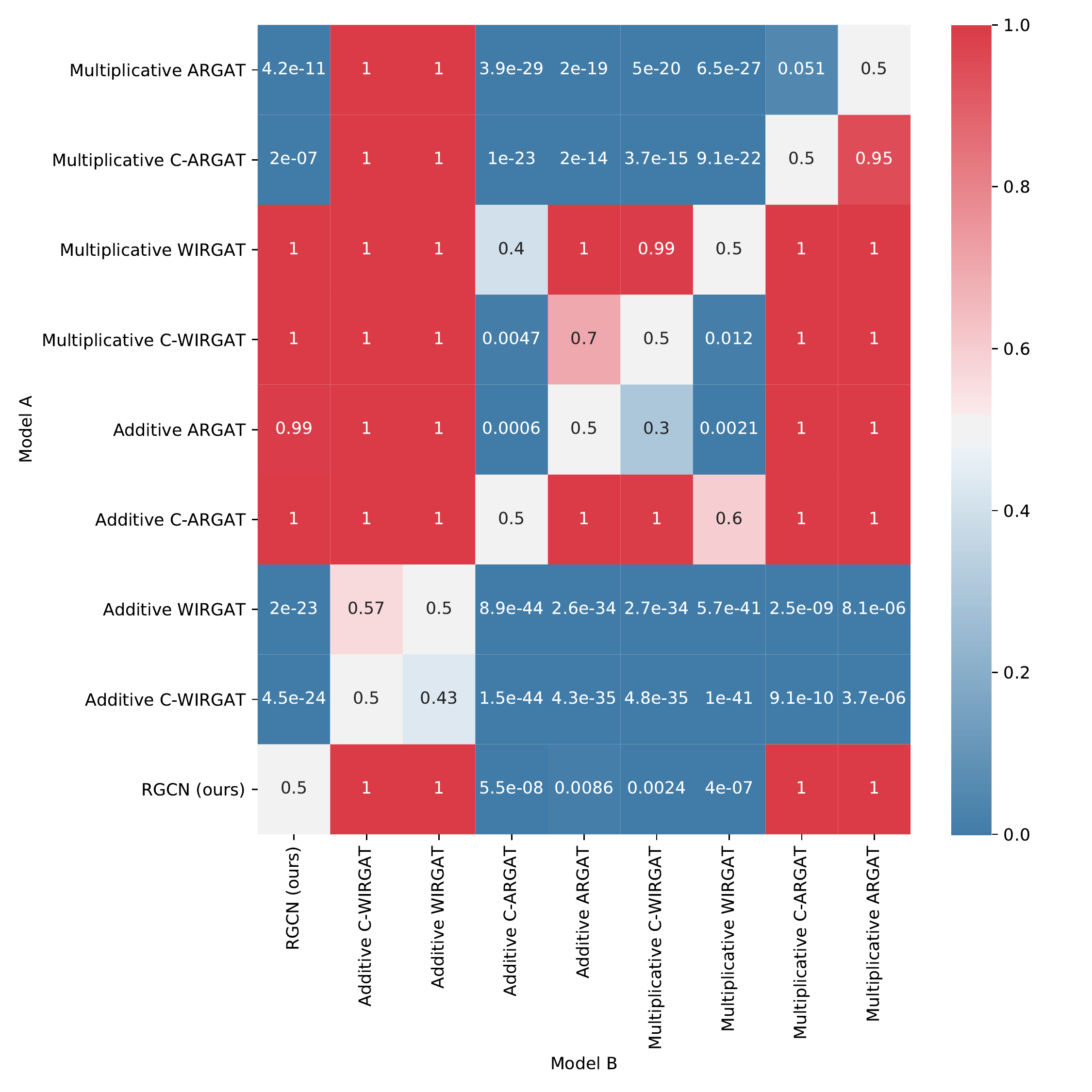}
  \caption{The $p$-values for Mann-Whitney $U$ test with alternative hypothesis $H_1$ of Model $A$ outperforming Model $B$ on AIFB.}
  \label{fig:aifb-p}
\end{figure}

\begin{figure}[ht]
  \centering
    \includegraphics[width=\textwidth]{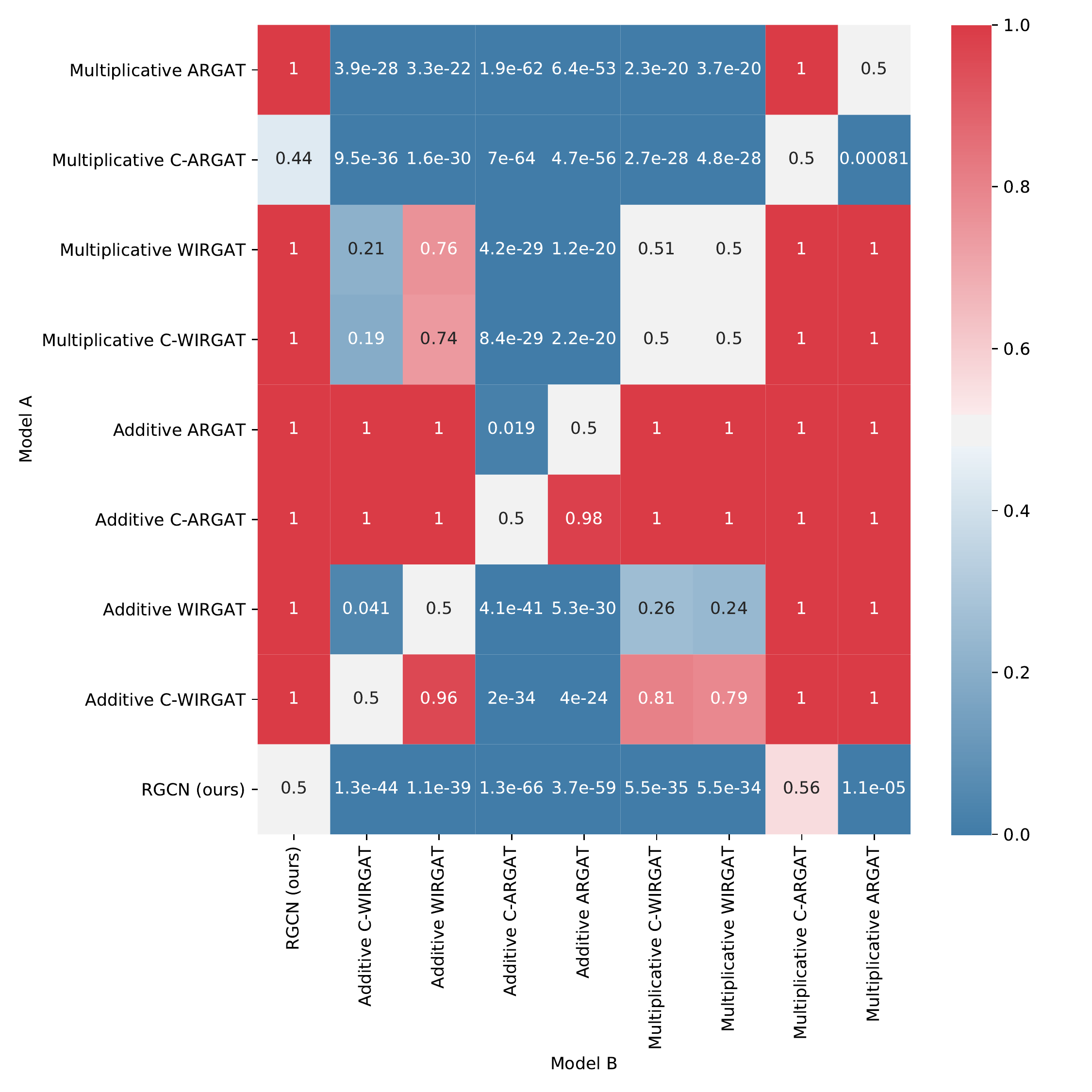}
  \caption{The $p$-values for Mann-Whitney $U$ test with alternative hypothesis $H_1$ of Model $A$ outperforming Model $B$ on MUTAG.}
  \label{fig:mutag-p}
\end{figure}

\begin{figure}[ht]
  \centering
    \includegraphics[width=\textwidth]{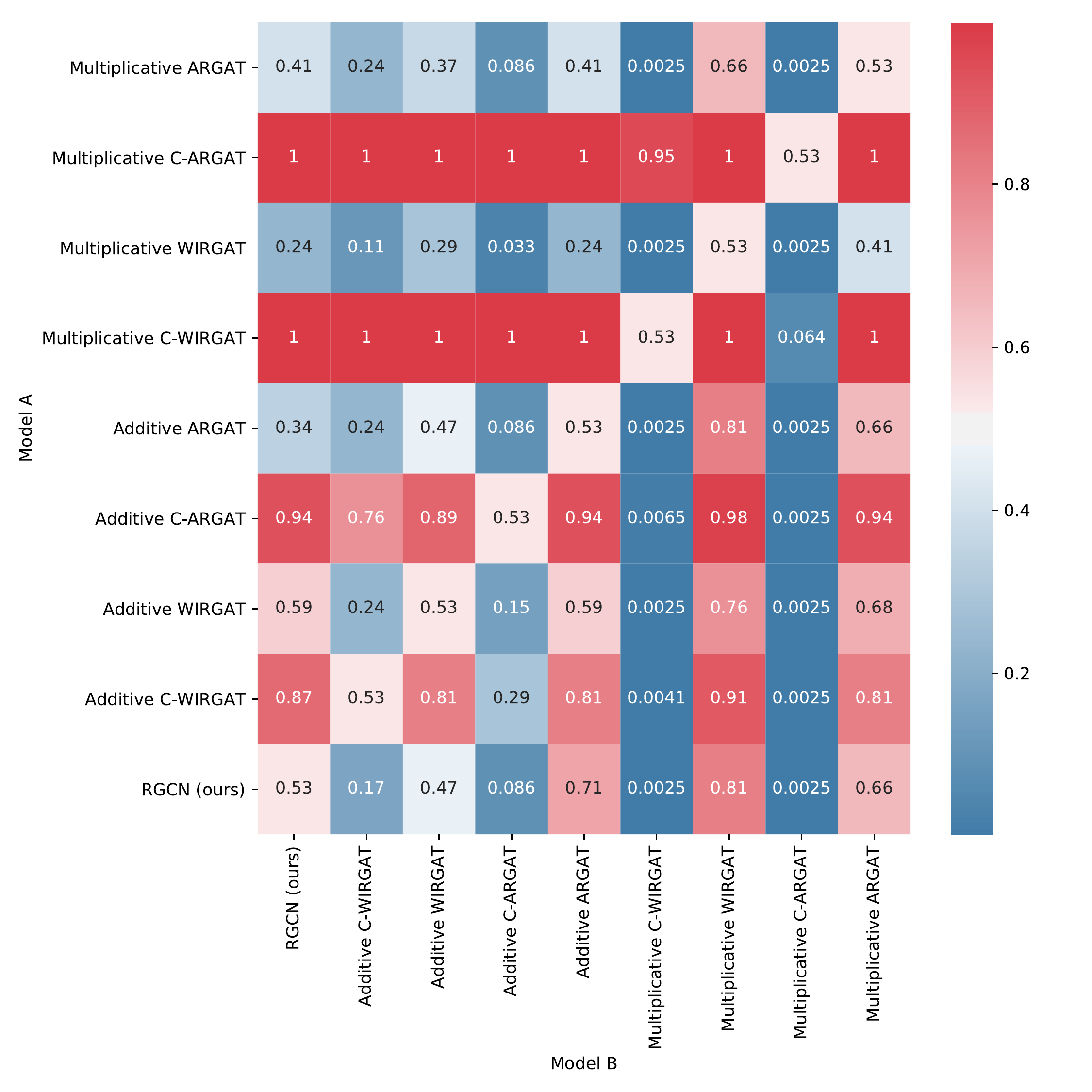}
  \caption{The $p$-values for Mann-Whitney $U$ test with alternative hypothesis $H_1$ of Model $A$ outperforming Model $B$ on TOX21.}
  \label{fig:tox-p}
\end{figure}

In order to determine if any of our model comparisons are significant, we employ the one-sided Mann-Whitney $U$ test
\cite{mann1947} as we are interested in the direction of movement (i.e. performance) and do not want to make any parametric assumptions about model response. For two populations $X$ and $Y$:
\begin{itemize}
  \item The null hypothesis $H_0$ is that the two populations are equal, and
  \item The alternative hypothesis $H_1$ is that the probability of an observation from population X exceeding an observation from population Y is larger than the probability of an observation from Y exceeding an observation from X; i.e., $H_1: P(X > Y) > P(Y > X)$.
\end{itemize}
We treat the empirical distributions of Model A as samples from population $X$ and the empirical distributions of Model B as samples from population $Y$. This allows us a window into whether, given a task, whether which is the better model out of a pair of models. Results on AIFB, MUTAG and TOX21 are given in \Cref{fig:aifb-p}, \Cref{fig:mutag-p} and \Cref{fig:tox-p} respectively.

\end{document}